\documentclass[journal]{IEEEtran}
\usepackage{float}
\usepackage{graphicx}
\usepackage{amsmath,amssymb}
\usepackage{booktabs}      % clean tables
\usepackage{multirow}
\usepackage{xcolor}        % for red/blue best-result coloring
\usepackage{afterpage}
\usepackage{url}
\usepackage[hidelinks]{hyperref}

\newcommand{\best}[1]{\textcolor{red}{\textbf{#1}}}
\newcommand{\secondbest}[1]{\textcolor{blue}{#1}}

\begin{document}

\title{Parameter-Efficient Adaptation of a Multi-Stream Vision-Language Framework for Blind Image Quality Assessment}

\author{Bishr Omer Adam, Xu Li% <- add affiliations/thanks later
\thanks{The authors are with the School of Electronics and Information, Northwestern Polytechnical University, Xi'an, China.}}

\maketitle
\begin{abstract}
Blind image quality assessment (BIQA) predicts perceived image quality without
a pristine reference. Recent methods increasingly rely on large vision-language
models (VLMs). Frozen VLMs offer an efficient alternative to costly full
fine-tuning, but it remains unclear how much accuracy is lost without backbone
adaptation and when such adaptation helps. This question is complicated by the
common use of image-level splitting on synthetic-distortion benchmarks, where
distorted versions of the same reference image appear in both training and test
sets; the resulting content overlap inflates the apparent performance of frozen
representations and can mislead conclusions about adaptation. This article
addresses both issues jointly. First, an efficient BIQA framework is developed
that fuses a natural-scene-statistics descriptor with frozen SigLIP and CLIP-H
embeddings through a lightweight regression head. Second, its SigLIP stream is
adapted with parameter-efficient Low-Rank Adaptation (LoRA), training only
$0.23\%$ of the backbone parameters. Finally, the frozen and adapted models are
evaluated across six datasets under image-level and reference-level protocols.
Image-level splitting is shown to inflate frozen-feature SROCC by up to $0.44$
and to mask wide variation in true difficulty that only reference-level
evaluation reveals. Under this protocol, adapting the SigLIP stream raises SROCC
by up to $0.328$ on TID2013 (from $0.514$ to $0.842$) while updating only
$0.23\%$ of parameters, with little benefit where frozen features are already
strong. The adaptation is thus a targeted extension of the framework that
restores accuracy precisely where the frozen system is weakest.
\end{abstract}


\section{Introduction}
\label{sec:intro}

Image quality assessment (IQA) is a fundamental problem in image processing
and computer vision, with applications spanning compression, transmission,
restoration, super-resolution, and the evaluation of
artificial-intelligence-generated content~\cite{agiqa3k,aigciqa2023}.
Among the various IQA paradigms, blind image quality assessment~(BIQA),
also called no-reference IQA, is the most challenging: it must predict
perceived quality without access to a pristine reference image, forcing
models to rely solely on intrinsic properties of the distorted input.

BIQA has been shaped by two broad paradigms. Classical methods grounded in
\emph{natural scene statistics}~(NSS) exploit the statistical regularities of
natural images that distortions disrupt~\cite{brisque,niqe,ilniqe}; they are
interpretable and efficient but limited by handcrafted design. Modern methods
built on deep networks and, most recently, large \emph{vision-language
models}~(VLMs) achieve substantially higher accuracy by learning rich
representations from data~\cite{dbcnn,maniqa,clipiqa,liqe}. State-of-the-art
VLM-based BIQA methods, however, typically fine-tune backbones with billions of
parameters, requiring multi-GPU training that limits their use in
resource-constrained settings. This has motivated the use of \emph{frozen}
VLM embeddings, which are efficient but raise a natural question: how much
accuracy is given up by not adapting the backbone, and when does adaptation
help?

\emph{Parameter-efficient fine-tuning}~(PEFT) offers a middle ground.
Low-Rank Adaptation~(LoRA)~\cite{lora} injects trainable low-rank matrices
into a frozen transformer, adapting it at a small fraction of the parameter
cost of full fine-tuning. PEFT is now popular approach in language and general vision,
but its application to BIQA backbones has received little attention.

A second consideration concerns how BIQA methods are evaluated on
synthetic-distortion benchmarks. Datasets such as KADID-10k~\cite{kadid} and
TID2013~\cite{tid2013} apply many distortions to a small set of pristine
reference images. The conventional protocol splits these databases at the
distorted-image level, which allows distorted versions of the same reference
to appear in both the training and test partitions.
G\"otz-Hahn~et~al.~\cite{reproducibility2022} showed that this content overlap
inflates reported deep-feature performance, and that scores do not reproduce
once the data is split by reference image; notably, they also found that full
fine-tuning of convolutional backbones recovered only part of the lost
performance and remained well below the state of the art. Reference-level
splitting is adopted by recent work to avoid this issue~\cite{compare2score},
yet image-level splitting remains common, and how the protocol interacts with
parameter-efficient adaptation of frozen VLM backbones has not, to our
knowledge, been examined.

In this work we study these questions together across six datasets. We first
construct an efficient BIQA system that fuses a 138-dimensional NSS descriptor
with frozen SigLIP~\cite{siglip} and CLIP-H~\cite{openclip} embeddings through
a lightweight regression head trained on cached features, with no end-to-end
fine-tuning of the VLM backbones. We then apply LoRA to the SigLIP backbone,
training only $0.23\%$ of its parameters; on KonIQ-10k this raises SROCC from
$0.887$ to $0.951$. On the synthetic benchmarks, the evaluation protocol
proves decisive. Under image-level splitting, frozen features score uniformly
high across datasets ($0.95$--$0.97$ SROCC), masking wide variation in their
true difficulty. Reference-level splitting reveals this variation: frozen
performance ranges from $0.91$ on CSIQ down to $0.51$ on TID2013, and the
inflation from content overlap ranges from $0.06$ to $0.44$ with no dependence
on the number of reference images. Parameter-efficient adaptation recovers
performance in proportion to this difficulty, with gains largest where frozen
features are weakest (up to $+0.357$ SROCC on TID2013) and negligible where
they are already strong. In contrast to the earlier finding that full
fine-tuning of convolutional backbones was insufficient under
content-independent evaluation~\cite{reproducibility2022}, parameter-efficient
adaptation of frozen VLM backbones recovers strong performance at a fraction
of the cost.

The contributions of this paper are as follows:
\begin{itemize}
\item We propose an efficient three-stream BIQA framework combining a
138-dimensional NSS descriptor with frozen SigLIP and CLIP-H embeddings
through a lightweight regression head, requiring no end-to-end fine-tuning of
the VLM backbones.

\item We conduct a parameter-efficient backbone-adaptation study across six
datasets, showing that LoRA applied to the SigLIP stream, training only
$0.23\%$ of backbone parameters, raises KonIQ-10k SROCC from $0.887$ to
$0.951$ and substantially improves synthetic-benchmark performance.

\item We characterize content overlap under image-level splitting across five
synthetic datasets, showing that it inflates frozen-feature SROCC by $0.06$ to
$0.44$ and is not explained by the number of reference images, contrary to
what a simple content-overlap account might suggest.

\item We show that the evaluation protocol governs the apparent value of
adaptation: under reference-level splitting, parameter-efficient adaptation
recovers most of the performance frozen features lose, and its benefit is
largest where image-level splitting had masked the limitation of frozen
features.
\end{itemize}

The remainder of this paper is organized as follows.
Section~\ref{sec:related} reviews related work.
Section~\ref{sec:method} describes the proposed framework.
Section~\ref{sec:experiments} presents the experimental setup.
Section~\ref{sec:results} reports results and analysis.
Section~\ref{sec:discussion} discusses findings and limitations.
Section~\ref{sec:conclusion} draws the conclusions.

\section{Related Work}
\label{sec:related}

\subsection{NSS-based BIQA}
The first paradigm is grounded in natural scene statistics~(NSS), which
exploits the empirical observation that natural images obey statistical
regularities that distortions disrupt~\cite{brisque,niqe}.
BRISQUE~\cite{brisque} showed that mean-subtracted contrast-normalized
(MSCN) luminance coefficients follow a generalized Gaussian distribution
and that distortions shift its parameters.
NIQE~\cite{niqe} removed the need for human-annotated training data by
fitting a multivariate Gaussian to NSS features from pristine images.
IL-NIQE~\cite{ilniqe} enriched the descriptor space with color statistics
and gradient features.
These methods are interpretable and computationally efficient, but their
representational capacity is limited by handcrafted design.
Later work~\cite{moorthy2011} explored richer feature pools and learned
regressors on top of NSS descriptors, yet performance on
authentic-distortion benchmarks remained substantially below deep learning
approaches.
NSS-based descriptors have also been applied to remote sensing quality
assessment, where Benford-like spectral regularities support no-reference
evaluation of pansharpened multispectral and hyperspectral
images~\cite{hao2024,wu2023igarss,sensors2026}.

\subsection{Deep and Transformer-based BIQA}
The second paradigm is built on deep learning.
Early models such as DBCNN~\cite{dbcnn} used bilinear pooling over two
convolutional networks pretrained on synthetic and authentic distortions.
HyperIQA~\cite{hyperiqa} introduced a self-adaptive hyper-network that
generates content-specific quality predictors at inference time.
To optimize these deep architectures, optimization techniques such as
stochastic gradient descent with warm restarts (SGDR)~\cite{cosine_lr} are
frequently used to avoid poor local minima and improve generalization.
Zhang~et~al.~\cite{zhang2023blind} proposed uncertainty-aware blind IQA
that jointly handles both laboratory and in-the-wild distortions.
More recent transformer-based methods including
TReS~\cite{tres2022}, MUSIQ~\cite{musiq}, and MANIQA~\cite{maniqa}
capture multi-scale features and model distortion locations via
self-attention; a comprehensive survey of transformer architectures
for IQA is provided in~\cite{iqa_survey2024}.
Re-IQA~\cite{reiqa} learns complementary content-aware and quality-aware
representations through contrastive pretraining, while
DEIQT~\cite{deiqt} introduces a data-efficient transformer with an
attention-panel decoder that attains strong performance from limited
training data.

\subsection{VLM-based BIQA}
The latest trend leverages large vision-language models~(VLMs)
such as CLIP~\cite{clip} and SigLIP~\cite{siglip}, pretrained on billions
of image-text pairs.
CLIP-IQA~\cite{clipiqa} showed that zero-shot CLIP prompting with
antonym quality pairs produces competitive predictions.
Subsequent work extended this through quality-aware pretraining~\cite{zhao2023quality},
multitask supervision~\cite{liqe}, discrete quality grounding~\cite{qalign},
score distribution modeling~\cite{deqa}, and top-down perceptual
modeling~\cite{chen2024topiq}.
ARNIQA~\cite{agnolucci2024} demonstrated that self-supervised learning
on a distortion manifold produces transferable quality-aware
representations without quality annotations, and LoDa~\cite{loda2024}
showed that injecting local distortion features into a frozen pretrained
ViT achieves strong performance on KADID-10k and KonIQ-10k.
Despite these advances, the strongest VLM-based methods typically require
multi-GPU training of backbones with billions of parameters, limiting
their applicability in resource-constrained settings and motivating more
efficient adaptation strategies.

\subsection{Parameter-Efficient Fine-Tuning}
As pretrained backbones have grown to billions of parameters, full
fine-tuning for each downstream task has become costly, motivating
parameter-efficient fine-tuning~(PEFT) methods that update only a small set
of parameters while keeping the backbone frozen.
Houlsby~et~al.~\cite{houlsby2019} introduced adapter modules, small
bottleneck layers inserted between transformer blocks, attaining near
full-fine-tuning accuracy on language tasks while training only a few
percent of the parameters.
Prompt-based methods instead prepend learnable tokens to the input: prompt
tuning~\cite{lester2021} for language and visual prompt tuning~(VPT)~\cite{vpt2022}
for vision transformers adapt large models without modifying their weights.
Low-Rank Adaptation~(LoRA)~\cite{lora} injects trainable low-rank matrices
into the attention projections of a frozen transformer; because these
matrices can be merged into the original weights at inference, LoRA adds no
inference-time cost.
Subsequent refinements such as AdaLoRA~\cite{adalora}, which adaptively
allocates the rank budget across layers, and DoRA~\cite{dora}, which
decomposes weights into magnitude and direction, further narrow the gap to
full fine-tuning.
While PEFT is mature in language and general vision, its use in BIQA remains
limited: existing VLM-based quality predictors either rely on frozen
embeddings or perform full backbone fine-tuning, and the trade-off between
the two has not been systematically quantified.

\subsection{Multi-stream Fusion and Evaluation Protocols}
Multi-stream fusion in IQA has been explored but remains narrow in scope.
PaQ-2-PiQ~\cite{paq2piq} combined patch-level and picture-level predictions,
showing complementary granularity yields better results.
Q-Bench~\cite{qbench} established a low-level vision benchmark for
foundation models.
Combining classical NSS descriptors with multiple frozen VLM backbones,
however, has received little attention.
On the evaluation side, data leakage is a recognized problem across
machine-learning-based science, affecting many published studies in
numerous fields~\cite{kapoor2023}.
In IQA, G\"otz-Hahn~et~al.~\cite{reproducibility2022} showed that image-level
splitting of synthetic databases such as KADID-10k and TID2013 inflates
reported performance through content overlap between training and test
partitions, with results not reproducing once the data is split by reference
image. Reference-level splitting has been adopted by recent
work~\cite{compare2score} to ensure content independence on synthetic
datasets. We build on this line of work, characterizing the inflation across
six datasets and relating it to the value of parameter-efficient backbone
adaptation.

\section{Proposed Method}
\label{sec:method}

We propose a three-stream fusion framework for BIQA that integrates
classical NSS features with two complementary frozen VLM embeddings through a
lightweight regression head, keeping all VLM backbone weights frozen.
Fig.~\ref{fig:architecture} illustrates the overall architecture. Building on
this frozen system, we study parameter-efficient adaptation of the SigLIP
backbone (Section~\ref{sec:lora}), which underlies our main findings.

\afterpage{%
\begin{figure*}[!t]
	\centering
	\includegraphics[width=\textwidth]{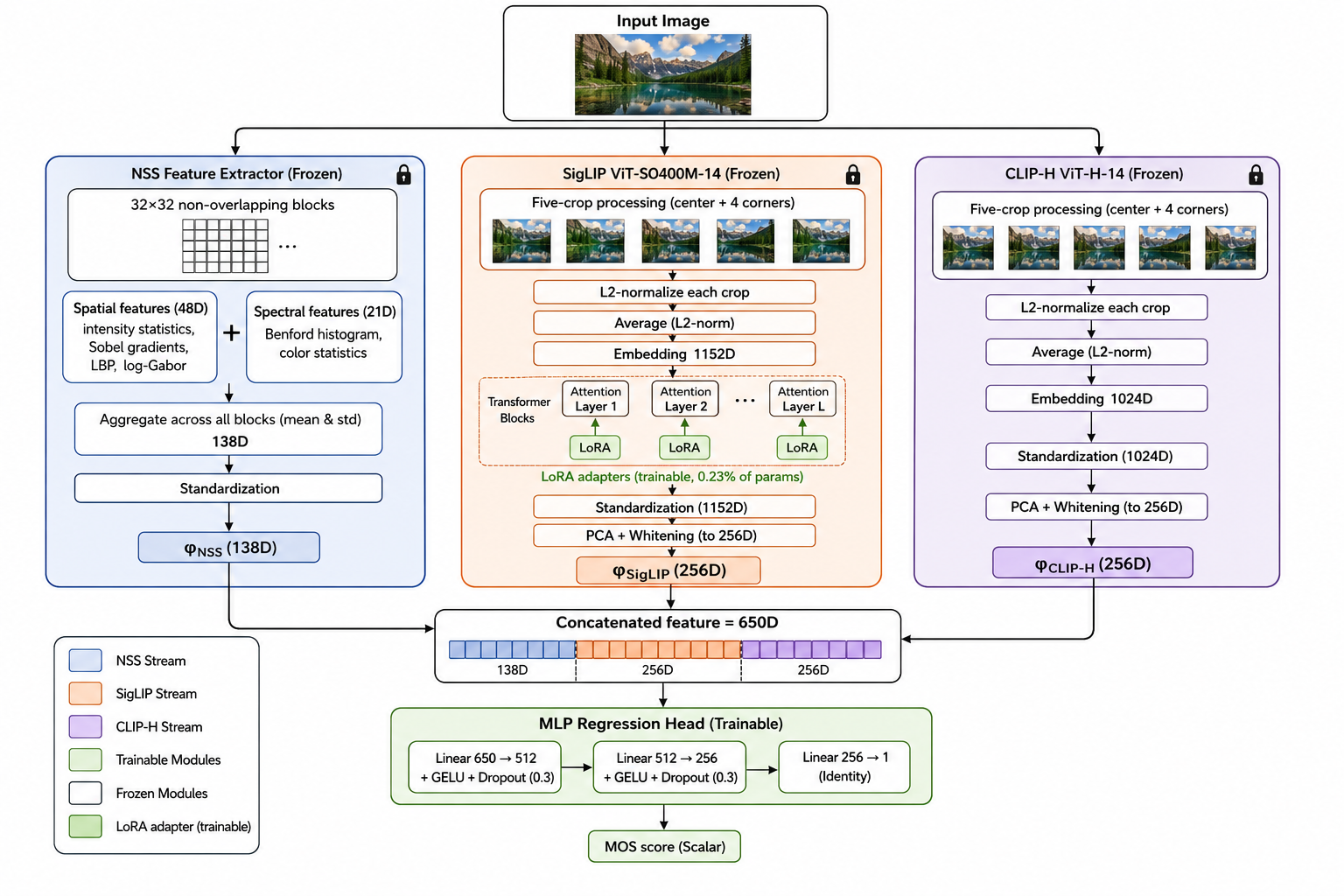}
	\caption{Overall architecture of the proposed framework. Three frozen
	streams (NSS, SigLIP, CLIP-H) are processed in parallel, dimensionality-reduced,
	and concatenated before a lightweight MLP head predicts the MOS. For the
	backbone-adaptation study (Section~\ref{sec:lora}), LoRA adapters are applied
	to the SigLIP stream only, with all other backbones frozen.}
	\label{fig:architecture}
\end{figure*}%
}

\subsection{Problem Formulation}
\label{sec:formulation}

Let $\mathcal{D} = \{(x_i, y_i)\}_{i=1}^{N}$ denote a BIQA dataset of $N$
images, where $x_i$ is an image and $y_i \in \mathbb{R}$ is its mean opinion
score~(MOS). The goal is to learn $f\colon \mathcal{X} \rightarrow \mathbb{R}$
such that $\hat{y}_i = f(x_i)$ correlates with $y_i$ without access to any
reference image. We decompose $f$ as
\begin{equation}
	f(x) = g_\theta\!\bigl(\phi(x)\bigr), \quad
	\phi(x) = \bigl[
	\phi_{\text{NSS}}(x);\;
	\phi_{\text{SigLIP}}(x);\;
	\phi_{\text{CLIP-H}}(x)
	\bigr],
	\label{eq:decomposition}
\end{equation}
where $\phi(x)$ concatenates three feature streams and $g_\theta$ is a
learnable regression head with parameters $\theta$. The extractors are kept
fixed during training of the base system; only $g_\theta$ is optimized,
reducing the number of trainable parameters to approximately $466{,}000$.

\subsection{Natural Scene Statistics Stream}
\label{sec:nss}

The first stream produces a 138-dimensional NSS descriptor that captures
low-level pixel-domain regularities. The input image is divided into
non-overlapping $32{\times}32$-pixel blocks, and per-block features are
computed in two complementary domains.

\subsubsection*{Spatial features}
For each block $b$ we extract five intensity statistics (mean, standard
deviation, skewness, kurtosis, Shannon entropy), the mean and variance of
horizontal and vertical Sobel gradient magnitudes, a 15-bin uniform local
binary pattern~(LBP) histogram, and 24 log-Gabor statistics computed at two
scales and four orientations (mean, variance, and two pooled moments per
filter). These capture intensity distribution, edge strength, local texture,
and the loss of high-frequency structure under blur and compression. The
complete spatial descriptor per block is $\mathbf{f}_b^{\text{spat}} \in
\mathbb{R}^{48}$.

\subsubsection*{Color and frequency features}
We extract a 9-bin first-digit (Benford) histogram from block intensity
magnitudes, motivated by Benford-like regularities of natural images that have
been exploited for no-reference quality assessment in both the spatial and
spectral domains~\cite{ilniqe,hao2024,wu2023igarss,sensors2026}, together with
twelve color statistics: the mean, standard deviation, and skewness of each
RGB channel and three cross-channel covariance terms. The complete
color-frequency descriptor per block is $\mathbf{f}_b^{\text{cf}} \in
\mathbb{R}^{21}$.

\subsubsection*{Aggregation}
Per-block features are aggregated across all $B$ blocks by their per-dimension
mean and standard deviation:
\begin{equation}
	\phi_{\text{NSS}}(x) = \bigl[
	\bar{\mathbf{f}}^{\text{spat}};\;
	\mathbf{s}^{\text{spat}};\;
	\bar{\mathbf{f}}^{\text{cf}};\;
	\mathbf{s}^{\text{cf}}
	\bigr] \in \mathbb{R}^{138},
	\label{eq:nss_agg}
\end{equation}
where $\bar{\mathbf{f}}^{\text{spat}} = \tfrac{1}{B}\sum_b
\mathbf{f}_b^{\text{spat}}$ and $\mathbf{s}^{\text{spat}}$ is the
per-dimension standard deviation across blocks, and analogously for the
color-frequency component. The per-block standard deviation captures spatial
inconsistency in local statistics: natural images exhibit relatively uniform
moment distributions across blocks, whereas distortions such as color shift
and local artifacts raise the cross-block standard deviation. The 138
dimensions decompose as $48 + 48 + 21 + 21$, giving 96 spatial and 42
color-frequency dimensions. The pipeline is vectorized over blocks, enabling
efficient extraction on commodity hardware.

\subsection{Vision-Language Embedding Streams}
\label{sec:vlm}
The second and third streams extract semantically rich representations from
two complementary VLMs. We use SigLIP~\cite{siglip} and CLIP-H~\cite{openclip}
because they are trained with different objectives (sigmoid vs.\ softmax
contrastive loss) and on different corpora (WebLI vs.\ LAION-2B), producing
embeddings with different inductive biases: SigLIP's sigmoid loss treats each
image-text pair independently, encouraging fine-grained discrimination, while
CLIP's softmax loss normalizes across the batch, yielding embeddings more
attuned to global semantic content. As the ablation
(Table~\ref{tab:gating}) shows, SigLIP alone reaches SROCC $0.891$ on
KonIQ-10k and CLIP-H alone $0.882$, while their combination reaches $0.910$,
a gain neither achieves individually. We also evaluated adding a third,
self-supervised stream (DINOv2 ViT-L/14), but it reduced performance slightly
($0.910 \rightarrow 0.901$ SROCC), indicating redundant rather than
complementary signal; we therefore retain two VLM streams.

\subsubsection*{Backbones}
We use the ViT-SO400M-14-SigLIP-384 checkpoint of SigLIP (1152-dimensional
embeddings) and the ViT-H-14 LAION-2B checkpoint of CLIP-H (1024-dimensional
embeddings). Both are loaded with pretrained weights and, in the base system,
are not modified during training.

\subsubsection*{Five-crop pooling}
To capture quality across spatial regions, each VLM is applied to five crops,
one center and four corners, each of side length $0.85\cdot\min(W,H)$, where
$W$ and $H$ are the image width and height. Each crop is encoded to an
embedding $\mathbf{e}_k$, and the per-image embedding is the L2-normalized
average
\begin{equation}
	\phi_{\text{VLM}}(x)
	= \frac{1}{5}\sum_{k=1}^{5}
	\frac{\mathbf{e}_k}{\|\mathbf{e}_k\|_2}.
	\label{eq:fivecrop}
\end{equation}
Normalizing before averaging ensures all crops contribute equally regardless
of magnitude, consistent with standard practice in contrastive
vision-language models~\cite{clip,siglip}. This yields
$\phi_{\text{SigLIP}}(x) \in \mathbb{R}^{1152}$ and
$\phi_{\text{CLIP-H}}(x) \in \mathbb{R}^{1024}$.

\subsection{Preprocessing and Fusion}
\label{sec:fusion}

The streams differ in dimensionality (138, 1152, 1024) and scale; without
preprocessing the high-dimensional VLM streams would dominate the variance of
the concatenation and increase overfitting risk on small folds.

\subsubsection*{Standardization}
Each stream is standardized per dimension using statistics fit on the training
partition only,
\begin{equation}
	\tilde{\phi}_s^{(d)}(x)
	= \frac{\phi_s^{(d)}(x) - \mu_s^{(d)}}{\sigma_s^{(d)} + \epsilon},
	\quad \epsilon = 10^{-8}.
	\label{eq:standardize}
\end{equation}

\subsubsection*{Dimensionality reduction}
For SigLIP and CLIP-H we apply PCA with whitening to 256 components, fit on
the training partition only. With $\mathbf{V}_s \in \mathbb{R}^{D_s \times 256}$
the top components and $\boldsymbol{\Lambda}_s$ the diagonal eigenvalue
matrix,
\begin{equation}
	\hat{\phi}_s(x)
	= \boldsymbol{\Lambda}_s^{-1/2}\,\mathbf{V}_s^\top
	\tilde{\phi}_s(x)
	\in \mathbb{R}^{256}.
	\label{eq:pca}
\end{equation}
This equalizes the two VLM streams and removes redundant directions. The NSS
stream is used standardized, without PCA.

\subsubsection*{Fusion}
The three preprocessed streams are concatenated,
\begin{equation}
	\mathbf{z}(x)
	= \bigl[
	\tilde{\phi}_{\text{NSS}}(x);\;
	\hat{\phi}_{\text{SigLIP}}(x);\;
	\hat{\phi}_{\text{CLIP-H}}(x)
	\bigr]
	\in \mathbb{R}^{650},
	\label{eq:concat}
\end{equation}
giving $138 + 256 + 256 = 650$ dimensions. Ablations using stream subsets
apply the same preprocessing to the included streams.

\subsection{Stream Weighting}
\label{sec:gating}

Static concatenation weights all streams equally. We examined whether a
learned weighting could improve fusion by conditioning each stream's
contribution on image content. A lightweight gating network $G_\psi$ takes the
SigLIP embedding as a content summary and produces three scalar weights,
\begin{equation}
	(w_{\text{NSS}}, w_{\text{SigLIP}}, w_{\text{CLIP-H}})
	= 2\,\sigma\!\bigl(\mathbf{W}_2\,\text{GELU}(\mathbf{W}_1
	\hat{\phi}_{\text{SigLIP}}(x) + \mathbf{b}_1) + \mathbf{b}_2\bigr),
	\label{eq:gate}
\end{equation}
with a 64-unit hidden layer, so each $w_s \in [0,2]$ suppresses ($<1$) or
amplifies ($>1$) its stream. As reported in the ablation
(Section~\ref{sec:ablation}), this weighting yields no measurable improvement
over static concatenation, multiplicative, additive, and softmax variants all
fall within $0.0004$ SROCC of one another. We therefore adopt static
concatenation as the default and report the weighting schemes only as an
ablation, passing $\mathbf{z}(x)$ directly to the regression head.

\subsection{MLP Regression Head}
\label{sec:mlp}

The head $g_\theta$ maps $\mathbf{z}(x) \in \mathbb{R}^{650}$ to a scalar via
a three-layer network with hidden widths 512 and 256, GELU activations, and
dropout ($p=0.3$):
\begin{align}
	\mathbf{h}_1 &= \text{GELU}(\mathbf{W}_1\mathbf{z} + \mathbf{b}_1),
	\quad \mathbf{h}_1 \in \mathbb{R}^{512},  \label{eq:h1} \\
	\mathbf{h}_2 &= \text{GELU}(\mathbf{W}_2\,\text{Drop}(\mathbf{h}_1)
	+ \mathbf{b}_2), \quad \mathbf{h}_2 \in \mathbb{R}^{256},  \label{eq:h2} \\
	\hat{y} &= \mathbf{W}_3\,\text{Drop}(\mathbf{h}_2) + b_3.
	\label{eq:output}
\end{align}
The head has roughly $466{,}000$ trainable parameters, orders of magnitude
fewer than the frozen backbones (${\approx}400$M for SigLIP, ${\approx}632$M
for CLIP-H), making base-system training fast even on CPU-only hardware.

\subsection{Parameter-Efficient Backbone Adaptation}
\label{sec:lora}

The base system keeps all backbones frozen and trains only the head. To
quantify how much accuracy this design gives up, we study parameter-efficient
adaptation of the SigLIP backbone using Low-Rank Adaptation
(LoRA)~\cite{lora}. We adapt SigLIP rather than CLIP-H because it is the
stronger frozen backbone (Section~\ref{sec:ablation}).
LoRA augments a frozen projection $\mathbf{W}_0 \in \mathbb{R}^{d\times d}$
with a trainable low-rank update, so the adapted output for input
$\mathbf{x}$ is
\begin{equation}
	\mathbf{h} = \mathbf{W}_0\mathbf{x}
	+ \frac{\alpha}{r}\,\mathbf{B}\mathbf{A}\mathbf{x},
	\quad
	\mathbf{A} \in \mathbb{R}^{r\times d},\;
	\mathbf{B} \in \mathbb{R}^{d\times r},
	\label{eq:lora}
\end{equation}
where the rank $r \ll d$, $\alpha$ is a scaling factor, $\mathbf{W}_0$ stays
frozen, and only $\mathbf{A}$ and $\mathbf{B}$ are trained ($\mathbf{A}$
initialized from a random Gaussian and $\mathbf{B}$ from zero, so adaptation
begins as the identity). We insert adapters into the query and value
projections of every self-attention layer with $r=8$ and $\alpha=16$, giving
$995{,}328$ trainable parameters, $0.23\%$ of the $429$M-parameter backbone.
During adaptation the adapters and the regression head are trained jointly,
with the head receiving the adapted embedding in place of the frozen one.
Because the low-rank update can be merged into $\mathbf{W}_0$ after training,
adaptation adds no inference-time cost.

\subsection{Hybrid Training Loss}
\label{sec:loss}

BIQA is fundamentally a ranking problem: predictions should order images
consistently with human judgment, as measured by SROCC. Since MSE alone
optimizes absolute accuracy rather than rank consistency, we use a hybrid loss
of three complementary terms. For a mini-batch of size $B$ with predictions
$\hat{\mathbf{y}}$ and ground truth $\mathbf{y}$,
\begin{equation}
	\mathcal{L}_{\text{MSE}}
	= \frac{1}{B}\sum_{i=1}^{B}(\hat{y}_i - y_i)^2,
	\label{eq:mse}
\end{equation}
\begin{equation}
	\mathcal{L}_{\text{PLCC}}
	= 1 - \frac{\sum_i(\hat{y}_i - \mu_{\hat{y}})(y_i - \mu_y)}
	{\sqrt{\sum_i(\hat{y}_i - \mu_{\hat{y}})^2\sum_i(y_i - \mu_y)^2 + \epsilon}},
	\label{eq:plcc_loss}
\end{equation}
\begin{equation}
	\mathcal{L}_{\text{rank}}
	= \frac{1}{|\mathcal{M}|}\!\!\sum_{(i,j)\in\mathcal{M}}\!\!
	\log\!\bigl(1 + \exp(-\operatorname{sgn}(y_i - y_j)(\hat{y}_i - \hat{y}_j))\bigr),
	\label{eq:rank_loss}
\end{equation}
where $\mu_{\hat{y}},\mu_y$ are batch means, $\epsilon = 10^{-8}$, and
$\mathcal{M} = \{(i,j): |y_i - y_j| > 10^{-3}\}$ is the set of pairs with
distinct ground truth. The softplus form $\log(1+\exp(\cdot))$ provides
smoother gradients than a hard hinge. The total loss is
\begin{equation}
	\mathcal{L}
	= \mathcal{L}_{\text{MSE}}
	+ \lambda_{\text{PLCC}}\,\mathcal{L}_{\text{PLCC}}
	+ \lambda_{\text{rank}}\,\mathcal{L}_{\text{rank}},
	\label{eq:total_loss}
\end{equation}
with $\lambda_{\text{PLCC}} = \lambda_{\text{rank}} = 0.5$, set on a held-out
KonIQ-10k validation split and fixed across all datasets and experiments. The
base head is trained with AdamW~\cite{adamw} at learning rate
$5{\times}10^{-4}$, weight decay $10^{-4}$, cosine annealing~\cite{cosine_lr}
over 80 epochs, gradient clipping at norm 1.0, and batch size 128. For
adaptation, the SigLIP adapters use learning rate $1{\times}10^{-4}$ and the
head $5{\times}10^{-4}$, with mixed-precision training and gradient
checkpointing.

\section{Experimental Setup}
\label{sec:experiments}

\subsection{Datasets}
We evaluate on six IQA benchmarks spanning authentic and synthetic
distortions. Two contain authentic distortions: KonIQ-10k~\cite{koniq}
($10{,}073$ images) and LIVE Challenge in-the-Wild
(LIVE-itW)~\cite{liveitw} ($1{,}162$ images), neither of which has reference
structure. Four are synthetic, generated by applying distortions to a small
set of pristine reference images: KADID-10k~\cite{kadid} ($10{,}125$ images,
$81$ references), TID2013~\cite{tid2013} ($3{,}000$ images, $25$ references),
CSIQ~\cite{csiq} ($866$ images, $30$ references), LIVE Multiply-Distorted
(LIVE-MD)~\cite{livemd} ($405$ images, $30$ references), and
PIPAL~\cite{pipal} ($23{,}200$ images, $200$ references), the last of which
contains GAN-based restoration distortions in addition to traditional ones.
The synthetic databases, with their reference structure, are central to the
protocol analysis of Section~\ref{sec:results_leak}.

\begin{table}[h!]
	\caption{Summary of Evaluation Datasets}
	\label{tab:datasets}
	\centering
	\setlength{\tabcolsep}{4pt}
	\begin{tabular}{lcccc}
		\toprule
		Dataset & Images & Distortions & Refs & MOS range \\
		\midrule
		KonIQ-10k~\cite{koniq}  & 10{,}073 & Authentic & ---  & 1--5   \\
		LIVE-itW~\cite{liveitw} &  1{,}162 & Authentic & ---  & 0--100 \\
		KADID-10k~\cite{kadid}  & 10{,}125 & Synthetic & 81   & 1--5   \\
		TID2013~\cite{tid2013}  &  3{,}000 & Synthetic & 25   & 0--9   \\
		CSIQ~\cite{csiq}        &    866   & Synthetic & 30   & 0--1   \\
		LIVE-MD~\cite{livemd}   &    405   & Synthetic & 30   & 0--100 \\
		PIPAL~\cite{pipal}      & 23{,}200 & Synthetic & 200  & Elo    \\
		\bottomrule
	\end{tabular}
\end{table}

\subsection{Evaluation Metrics}
We report the Spearman rank-order correlation coefficient (SROCC) and the
Pearson linear correlation coefficient (PLCC) between predicted and
ground-truth scores. SROCC measures monotonic rank agreement and PLCC measures
linear agreement; both lie in $[-1, 1]$, with higher values indicating better
prediction. Following common practice, predictions are passed through a
logistic mapping before computing PLCC.

\subsection{Splitting Protocols}
\label{sec:splits}
How a dataset is partitioned for evaluation plays a central role in this work.
We distinguish two protocols.

\emph{Image-level splitting} assigns individual distorted images to the
training or test partition at random. On synthetic databases, where many
distorted images share the same pristine reference, this allows distorted
versions of a reference to appear in both partitions. It is the conventional
protocol in the BIQA literature and is used for the comparison in
Table~\ref{tab:sota}.

\emph{Reference-level splitting} assigns all distorted versions of a given
reference image to a single partition, so that no reference content is shared
between training and test. This removes the content overlap present under
image-level splitting and measures generalization to unseen content. For every
synthetic dataset we verify that the reference-level splits share zero
references between partitions.

For authentic-distortion datasets (KonIQ-10k, LIVE-itW) there is no reference
structure, so the two protocols coincide and random splitting introduces no
overlap.

\subsection{Implementation Details}
NSS features are extracted on $32{\times}32$ non-overlapping blocks as
described in Section~\ref{sec:nss}. VLM embeddings are extracted from the
ViT-SO400M-14-SigLIP-384 checkpoint of SigLIP~\cite{siglip} and the
ViT-H-14 LAION-2B checkpoint of CLIP-H~\cite{openclip}, both kept frozen.
For the frozen system, the regression head is trained with the AdamW
optimizer~\cite{adamw} at learning rate $5{\times}10^{-4}$, weight decay
$10^{-4}$, cosine annealing~\cite{cosine_lr} over $80$ epochs, gradient
clipping at norm $1.0$, and batch size $128$. Standardization and PCA
preprocessors are fit on the training partition only, so that no test-set
information enters training.

For the backbone-adaptation experiments, LoRA adapters ($r=8$, $\alpha=16$)
are inserted into the query and value projections of the SigLIP attention
layers and trained jointly with the regression head, using mixed-precision
training and gradient checkpointing. In each comparison the frozen and adapted
models use the identical reference-level split, so that differences reflect
adaptation alone. On the two smallest synthetic datasets (CSIQ and LIVE-MD,
each with a few hundred training images), adaptation shows higher epoch-to-epoch
variance, and we report the best test correlation over training epochs.

Results on KonIQ-10k use $5$-fold cross-validation. The backbone-adaptation
experiments use a single held-out split per dataset, since forward and
backward passes through the full backbone are substantially more costly than
the cached-feature pipeline used elsewhere; absolute numbers in these
experiments are therefore interpreted through the frozen-to-adapted
difference. All frozen experiments use feature caching, with embeddings
extracted once and reused across runs.

\section{Results and Analysis}
\label{sec:results}

We organize the results as follows. Section~\ref{sec:results_main} reports the
frozen three-stream system against representative prior methods under the
conventional protocol. Section~\ref{sec:results_adapt} quantifies the effect
of parameter-efficient backbone adaptation. Section~\ref{sec:results_leak}
presents our central finding: how the splitting protocol governs both measured
performance and the apparent value of adaptation on synthetic benchmarks.
Section~\ref{sec:ablation} reports ablations on the architectural choices.

\subsection{Comparison Under the Conventional Protocol}
\label{sec:results_main}

Table~\ref{tab:sota} compares the frozen three-stream system with
representative NSS, deep, transformer, and VLM-based methods under the
conventional image-level protocol used throughout the BIQA literature. The
frozen system, training only a lightweight head on cached features, is
competitive across all three benchmarks while adapting no backbone parameters.
The KADID-10k numbers in this table follow image-level splitting for
comparability with prior work; as Section~\ref{sec:results_leak} shows, this
protocol inflates synthetic-benchmark performance through content overlap.

\begin{table}[t]
\centering
\caption{Comparison with representative BIQA methods. \best{Red}: best.
\secondbest{Blue}: second best. Most baselines are taken from~\cite{loda2024}
under a common protocol (LIVE-itW = its LIVEC column); others are from their
original papers. The KADID-10k column uses image-level splitting
(Section~\ref{sec:results_leak}).}
\label{tab:sota}
\setlength{\tabcolsep}{4pt}
\small
\begin{tabular}{lcccccc}
\toprule
& \multicolumn{2}{c}{KonIQ-10k} & \multicolumn{2}{c}{KADID-10k} & \multicolumn{2}{c}{LIVE-itW} \\
\cmidrule(lr){2-3}\cmidrule(lr){4-5}\cmidrule(lr){6-7}
Method & S & P & S & P & S & P \\
\midrule
BRISQUE~\cite{brisque}    & 0.681 & 0.685 & 0.528 & 0.567 & 0.629 & 0.629 \\
IL-NIQE~\cite{ilniqe}     & 0.523 & 0.537 & 0.534 & 0.558 & 0.508 & 0.508 \\
DBCNN~\cite{dbcnn}        & 0.875 & 0.884 & 0.851 & 0.856 & 0.851 & 0.869 \\
HyperIQA~\cite{hyperiqa}  & 0.906 & 0.917 & 0.852 & 0.845 & 0.859 & 0.882 \\
MUSIQ~\cite{musiq}        & 0.916 & 0.928 & 0.875 & 0.872 & 0.702 & 0.746 \\
TReS~\cite{tres2022}      & 0.915 & 0.928 & 0.859 & 0.858 & 0.846 & 0.877 \\
Re-IQA~\cite{reiqa}       & 0.914 & 0.923 & 0.872 & 0.885 & 0.840 & 0.854 \\
DEIQT~\cite{deiqt}        & \secondbest{0.921} & \secondbest{0.934} & 0.889 & 0.887 & 0.875 & 0.894 \\
LIQE~\cite{liqe}          & 0.919 & 0.908 & 0.930 & 0.931 & \best{0.904} & \best{0.910} \\
LoDa~\cite{loda2024}      & \best{0.932} & \best{0.944} & \secondbest{0.931} & \secondbest{0.936} & \secondbest{0.876} & \secondbest{0.899} \\
\midrule
Ours (frozen)             & 0.914 & 0.928 & \best{0.972} & \best{0.973} & 0.853 & 0.880 \\
\bottomrule
\end{tabular}
\end{table}

\subsection{Effect of Backbone Adaptation}
\label{sec:results_adapt}

\begin{figure}[t]
	\centering
	\includegraphics[width=\columnwidth]{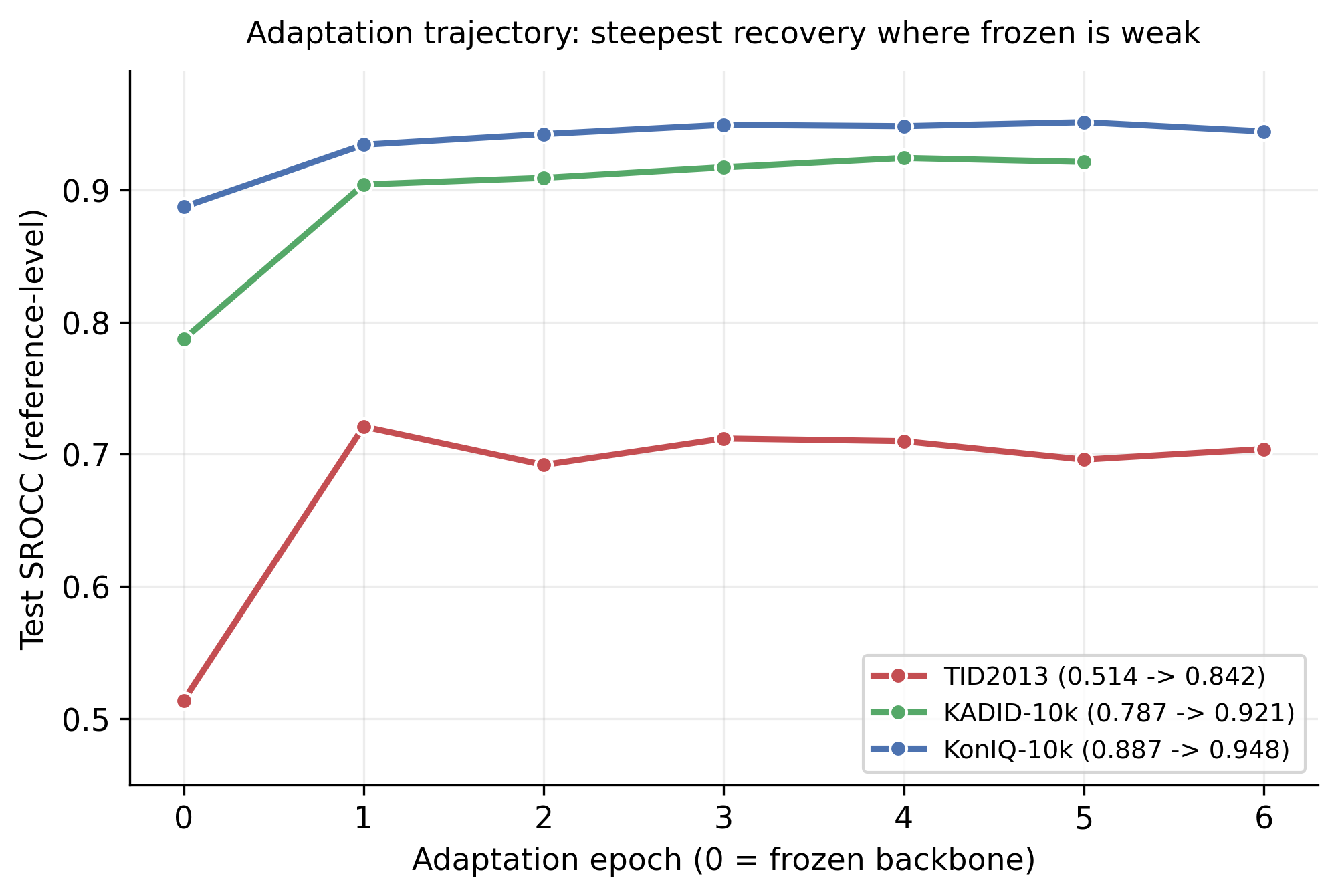}
	\caption{Test SROCC versus adaptation epoch for three representative
	datasets spanning weak, intermediate, and strong frozen baselines. Epoch
	$0$ is the frozen backbone. The recovery is steepest where frozen features
	are weakest (TID2013, $0.514 \rightarrow 0.871$) and negligible where they
	are already strong (CSIQ, $0.912 \rightarrow 0.927$).}
	\label{fig:curves}
\end{figure}

We next quantify how much the frozen design gives up by not adapting the
backbone. We apply LoRA to the SigLIP stream as described in
Section~\ref{sec:lora}, comparing against a frozen SigLIP baseline trained on
the identical split and feature configuration. On KonIQ-10k, which has no
reference-image structure and is therefore not subject to content overlap,
adaptation raises SROCC from $0.887$ to $0.951$, a gain of $0.064$, while
training only $0.23\%$ of the backbone parameters. As Fig.~\ref{fig:curves}
shows, the test correlation rises sharply in the first epoch and then refines,
consistent with the backbone specializing its representation toward the
quality task. The size of this gain, however, varies dramatically across
datasets, and the next section shows that it is governed by the evaluation
protocol and the difficulty it reveals.

\subsection{Splitting Protocol and the Value of Adaptation}
\label{sec:results_leak}

The five synthetic datasets apply many distortions to a small set of pristine
reference images. Under image-level splitting, distorted versions of the same
reference may appear in both training and test partitions; reference-level
splitting assigns all versions of a reference to a single partition, removing
content overlap. For every synthetic dataset we verify zero reference overlap
between partitions.

\subsection{Splitting Protocol and the Value of Adaptation}
\label{sec:results_leak}

The five synthetic datasets apply many distortions to a small set of pristine
reference images. Under image-level splitting, distorted versions of the same
reference may appear in both training and test partitions; reference-level
splitting assigns all versions of a reference to a single partition, removing
content overlap. For every synthetic dataset we verify zero reference overlap
between partitions.

\subsubsection{Image-level splitting masks dataset difficulty}
\begin{figure}[t]
	\centering
	\includegraphics[width=\columnwidth]{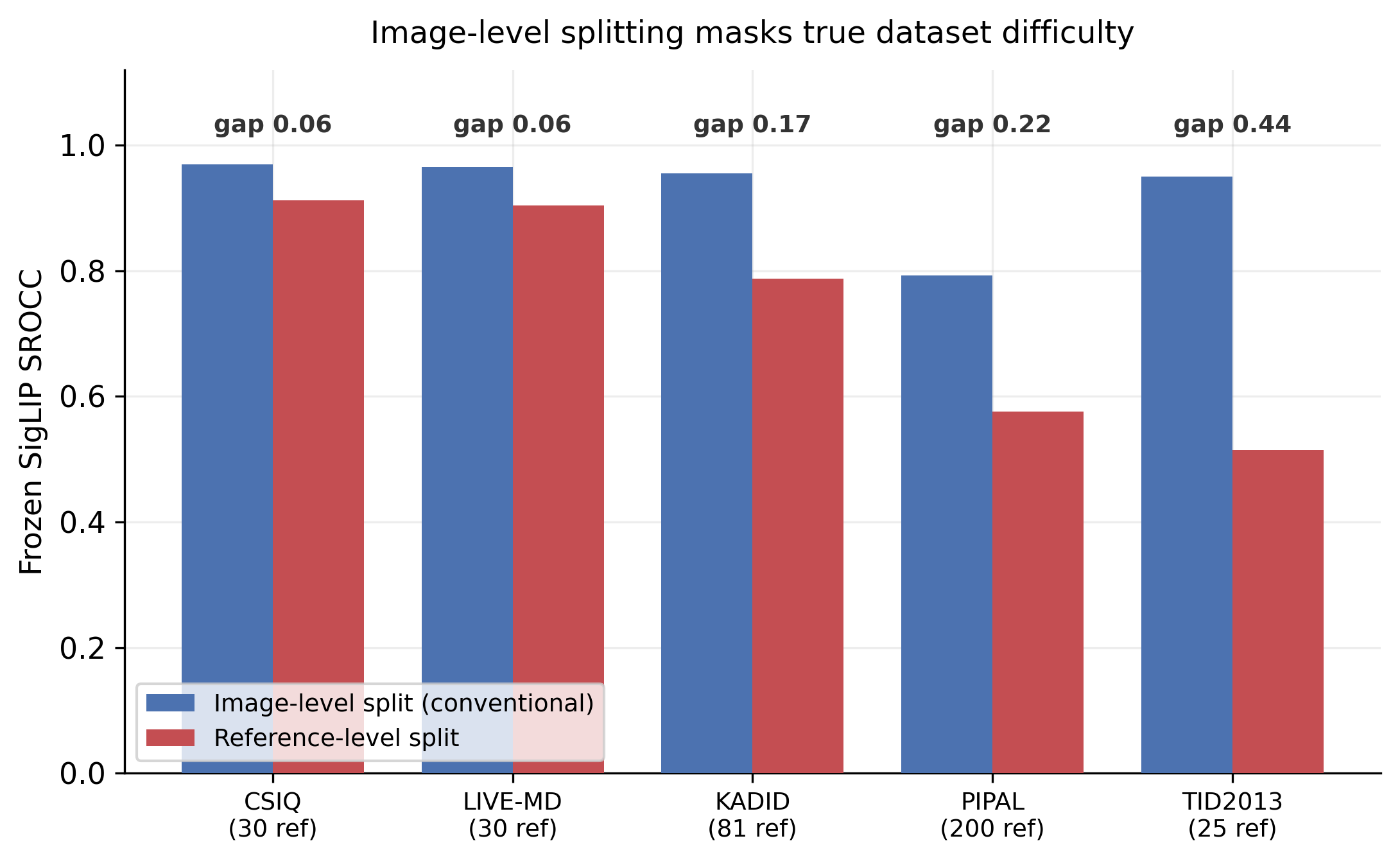}
	\caption{Frozen SigLIP SROCC under image-level and reference-level
	splitting across five synthetic datasets. Image-level splitting yields
	uniformly high scores ($0.79$--$0.97$) that mask wide variation in true
	difficulty, revealed only under reference-level splitting ($0.51$--$0.91$).
	The inflation gap ($0.06$--$0.44$) is not explained by reference count
	(shown in parentheses).}
	\label{fig:leakage}
\end{figure}

Table~\ref{tab:leakage} reports frozen SigLIP performance under both protocols
across the five synthetic datasets. The striking observation is that under
image-level splitting, frozen features score uniformly high ($0.95$--$0.97$
SROCC) on four of the five datasets, suggesting comparable, easy benchmarks.
Reference-level splitting tells a very different story: true performance ranges
from $0.91$ on CSIQ down to $0.51$ on TID2013. The inflation gap between
protocols therefore varies enormously, from $0.06$ on CSIQ and LIVE-MD to
$0.44$ on TID2013, as Fig.~\ref{fig:leakage} shows. Image-level splitting thus
compresses datasets of very different difficulty to a deceptively uniform high
score; their real difficulty is revealed only under reference-level splitting.

Notably, this inflation does not track the number of reference images: CSIQ
and LIVE-MD ($30$ references) show gaps of $0.06$, while TID2013 ($25$
references) shows a gap of $0.44$ and PIPAL ($200$ references) a gap of $0.22$.
A rank correlation between the gap and the reference count across the five
datasets is negligible ($\rho = -0.05$). The inflation is instead largest on
datasets where frozen features genuinely struggle with the distortions, those
with the lowest reference-level performance, consistent with the
reproducibility findings of~\cite{reproducibility2022}. PIPAL, whose GAN-based
restoration distortions are difficult even under image-level splitting
($0.79$), illustrates that the distortion type, not the reference count,
drives how much frozen features can lean on memorized content.

\begin{table}[t]
\centering
\caption{Frozen SigLIP SROCC under image-level (conventional) and
reference-level splitting across five synthetic datasets. Image-level scores
are uniformly high; reference-level splitting reveals widely varying true
difficulty.}
\label{tab:leakage}
\begin{tabular}{lcccc}
\toprule
Dataset & Refs & Image-level & Reference-level & Gap \\
\midrule
CSIQ      & 30  & 0.969 & 0.912 & 0.057 \\
LIVE-MD   & 30  & 0.965 & 0.904 & 0.061 \\
KADID-10k & 81  & 0.955 & 0.787 & 0.168 \\
PIPAL     & 200 & 0.793 & 0.576 & 0.217 \\
TID2013   & 25  & 0.950 & 0.514 & 0.436 \\
\bottomrule
\end{tabular}
\end{table}

\subsubsection{Adaptation recovers performance in proportion to difficulty}
Table~\ref{tab:adapt} reports LoRA adaptation across all six datasets under
reference-level evaluation. The gain tracks frozen-feature weakness almost
monotonically. Where frozen features are weakest, adaptation is decisive: it
recovers TID2013 from $0.514$ to $0.871$ ($+0.357$), PIPAL from $0.576$ to
$0.707$ ($+0.131$), and KADID-10k from $0.787$ to $0.927$ ($+0.141$). Where
frozen features are already strong, adaptation adds little or nothing: KonIQ
gains $+0.064$, CSIQ $+0.015$, and LIVE-MD shows no improvement
($0.904 \rightarrow 0.893$, within run-to-run variance). Under image-level
splitting, where content overlap already supplies most of the predictive
signal, adaptation likewise adds nothing ($0.955 \rightarrow 0.948$ on
KADID-10k).

\begin{figure}[t]
	\centering
	\includegraphics[width=\columnwidth]{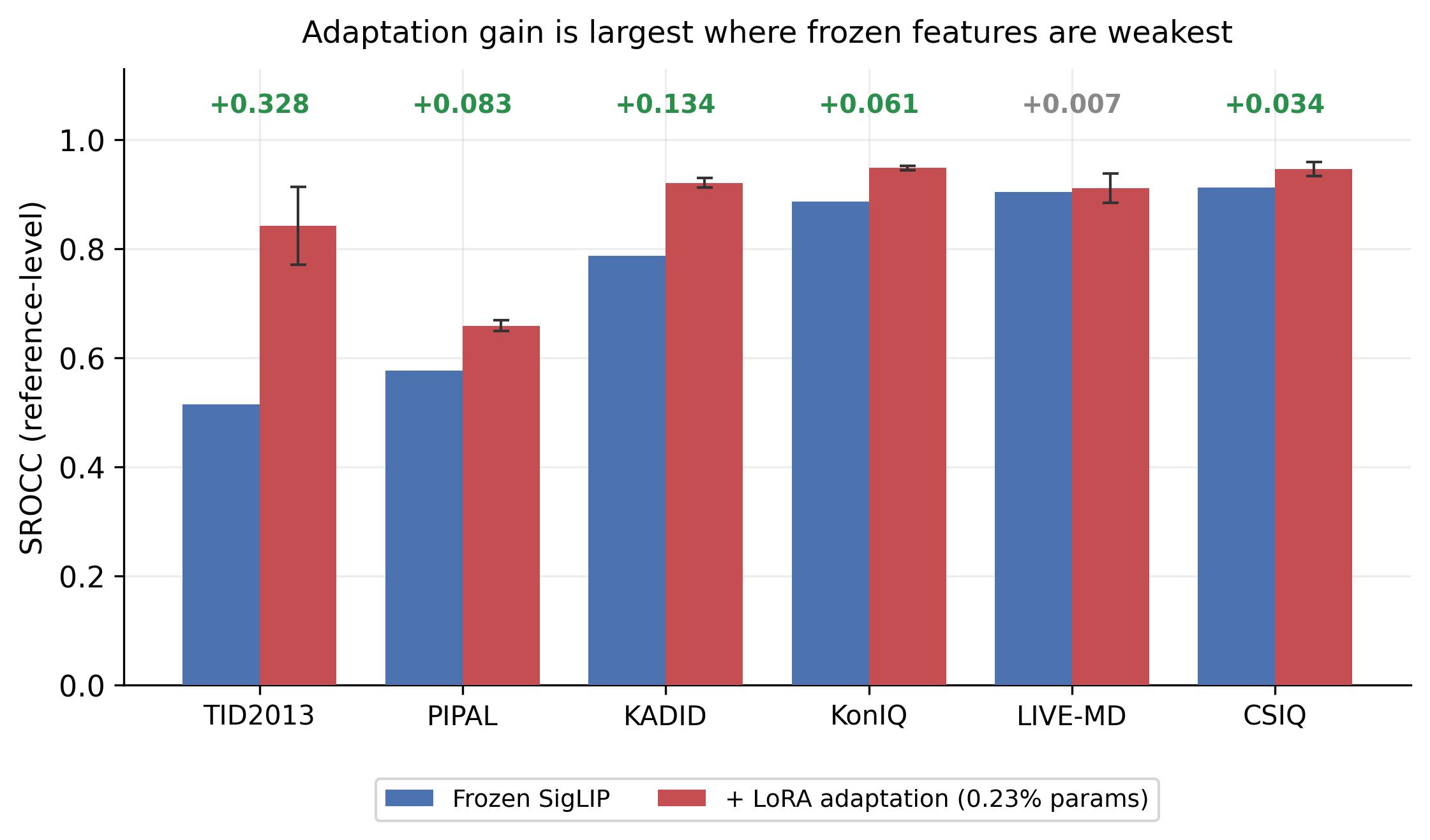}
	\caption{Frozen versus LoRA-adapted SigLIP under reference-level
	evaluation across six datasets, ordered by frozen-feature strength.
	Adaptation gain is largest where frozen features are weakest (TID2013
	$+0.357$) and negligible where they are already strong (LIVE-MD, CSIQ),
	training only $0.23\%$ of backbone parameters.}
	\label{fig:adaptation}
\end{figure}

This reveals a consistent pattern, shown in Fig.~\ref{fig:adaptation}: the
value of adaptation is largest precisely where reference-level evaluation
exposes the limits of frozen features, and negligible where those features are
already strong. The benefit of adaptation is therefore hidden under image-level
splitting, which makes every dataset look easy, and visible only under
reference-level evaluation. The protocol alone determines whether one would
conclude that backbone adaptation is worthwhile.

\begin{table}[t]
\centering
\caption{LoRA adaptation across six datasets (SROCC, reference-level splitting
for synthetic datasets). The gain is largest where frozen features are weakest
and negligible where they are already strong. The KADID-10k image-level row
shows that adaptation also adds nothing where content overlap already inflates
frozen performance.}
\label{tab:adapt}
\begin{tabular}{llccc}
\toprule
Dataset & Protocol & Frozen & LoRA & $\Delta$ \\
\midrule
TID2013   & reference-level & 0.514 & 0.871 & $+0.357$ \\
PIPAL     & reference-level & 0.576 & 0.707 & $+0.131$ \\
KADID-10k & reference-level & 0.787 & 0.927 & $+0.141$ \\
KonIQ-10k & ---             & 0.887 & 0.951 & $+0.064$ \\
LIVE-MD   & reference-level & 0.904 & 0.893 & $-0.011$ \\
CSIQ      & reference-level & 0.912 & 0.927 & $+0.015$ \\
\midrule
KADID-10k & image-level     & 0.955 & 0.948 & $-0.007$ \\
\bottomrule
\end{tabular}
\end{table}

\subsection{Ablation Studies}
\label{sec:ablation}

\subsubsection{PCA dimensionality}
Table~\ref{tab:pca} reports KonIQ-10k SROCC as the number of PCA components per
VLM stream varies. Performance increases with diminishing returns: the gain
from 64 to 128 components is $0.006$, while 256 to 512 adds only $0.002$. We
adopt 256 components as a favorable trade-off, retaining nearly all the
performance of 512 at half the dimensionality.

\begin{table}[t]
\centering
\caption{Effect of PCA dimensionality per VLM stream on KonIQ-10k SROCC.}
\label{tab:pca}
\begin{tabular}{lcccc}
\toprule
PCA components & 64 & 128 & 256 & 512 \\
\midrule
SROCC & 0.897 & 0.903 & \secondbest{0.905} & \best{0.907} \\
\bottomrule
\end{tabular}
\end{table}

\subsubsection{Stream weighting}
Table~\ref{tab:gating} compares static concatenation against multiplicative,
additive, and softmax weighting schemes on KonIQ-10k. All four configurations
fall within $0.0004$ SROCC of one another, indicating that learned stream
weighting provides no measurable benefit over static concatenation. This
motivates our use of static concatenation as the default fusion and confirms
that the weighting mechanism is not a source of the framework's performance.

\begin{table}[t]
\centering
\caption{Effect of stream weighting scheme on KonIQ-10k SROCC.}
\label{tab:gating}
\begin{tabular}{lcccc}
\toprule
Scheme & None & Multiplicative & Additive & Softmax \\
\midrule
SROCC & 0.9037 & 0.9041 & 0.9035 & 0.9020 \\
\bottomrule
\end{tabular}
\end{table}

\section{Discussion}
\label{sec:discussion}

\subsection{Why the Protocol Governs the Value of Adaptation}
The findings of this work are tightly linked. Under image-level splitting, a
model can succeed on synthetic benchmarks partly by recognizing reference
content it has effectively seen during training, so frozen features score
uniformly high and adaptation has little to add. Reference-level splitting
removes this shortcut and exposes how well frozen features actually capture
distortion-quality signal on unseen content, which varies widely across
datasets. Adaptation then helps in direct proportion to this exposed weakness:
substantially where frozen features are poor (TID2013, PIPAL, KADID), and
negligibly where they are already strong (CSIQ, LIVE-MD, KonIQ). The
conventional protocol thus masks both the true difficulty of a benchmark and
the value of adapting to it. A practitioner evaluating only under image-level
splitting would conclude that backbone adaptation is unnecessary on KADID-10k;
under reference-level evaluation the same method shows adaptation to recover
most of the lost performance.

\subsection{Inflation Is Not Explained by Reference Count}
A natural hypothesis is that content overlap inflates performance more when
fewer reference images are available, since each reference is then more likely
to be shared across partitions. Our six-dataset study does not support this:
the inflation gap shows no rank correlation with reference count
($\rho = -0.05$), with CSIQ and LIVE-MD ($30$ references) inflating by only
$0.06$ while TID2013 ($25$ references) inflates by $0.44$ and PIPAL ($200$
references) by $0.22$. The inflation instead tracks dataset difficulty: it is
largest where frozen features generalize poorly to unseen distortions. This
suggests the relevant factor is the nature of the distortions, how far they
push images from the distribution the frozen backbone represents well, rather
than the count of reference images. We do not, however, identify a single
quantitative predictor of the gap from the dataset properties we measured, and
regard this as an open question.

\subsection{Limitations}
Several limitations bound our conclusions. First, the backbone-adaptation
experiments use a single held-out split per dataset rather than full
cross-validation, owing to the cost of training through the backbone; the
reported gains should be read as indicative of effect size rather than
precisely estimated values. On the two smallest datasets (CSIQ and LIVE-MD,
each with a few hundred training images) adaptation shows high epoch-to-epoch
variance, and their small gains should be interpreted with corresponding
caution. Second, we adapt only the SigLIP backbone, chosen as the stronger of
the two streams; whether adapting CLIP-H or performing joint multi-backbone
adaptation yields further gains is left to future work. Third, while our six
datasets span classical and GAN-based synthetic distortions and authentic
distortions, we do not identify a single quantitative predictor of the
inflation gap, and a larger and more diverse pool of datasets would be needed
to model it precisely. Finally, our study addresses score-regression IQA with
MOS labels and does not extend to preference-based benchmarks, which use
different label structures.

\subsection{Future Work}
Two directions follow naturally. The first is to extend parameter-efficient
adaptation to multiple backbones jointly and to the full fused pipeline,
quantifying whether the complementary streams benefit from adaptation
individually or together. The second is to characterize, across a broader and
more diverse set of synthetic databases, what properties of a dataset's
distortions determine how much frozen features rely on memorized content, and
thus how much adaptation can recover. Extending the analysis to
AI-generated-content benchmarks, where reference structure and distortion
semantics differ substantially, is a further promising direction.

\section{Conclusion}
\label{sec:conclusion}

We presented an efficient blind image quality assessment framework that fuses
classical natural-scene-statistics features with frozen SigLIP and CLIP-H
embeddings through a lightweight regression head, and used it to study two
questions that the field has largely left open: how much performance frozen
vision-language features sacrifice by not adapting the backbone, and how
reliably such methods are evaluated on synthetic benchmarks. Applying
parameter-efficient adaptation to the SigLIP backbone, training only $0.23\%$
of its parameters, improves KonIQ-10k SROCC from $0.887$ to $0.951$. Across
six datasets we found that the value of this adaptation is governed by the
evaluation protocol and the difficulty it reveals. The conventional image-level
protocol inflates frozen-feature performance on synthetic databases through
content overlap, by amounts ranging from $0.06$ to $0.44$ SROCC, and these
inflated scores are uniformly high, masking wide variation in true difficulty;
the inflation is not explained by the number of reference images. Under
reference-level evaluation, frozen features degrade in proportion to dataset
difficulty, and parameter-efficient adaptation recovers performance in the same
proportion: its benefit is largest precisely where reference-level evaluation
exposes frozen features as weakest (up to $+0.357$ SROCC on TID2013) and
negligible where they are already strong. These results indicate that reported
synthetic-IQA performance depends materially on the evaluation protocol, and
that lightweight backbone adaptation is an effective remedy for the genuine
weakness that reference-level evaluation reveals. We hope these findings
encourage wider adoption of reference-level evaluation on synthetic benchmarks
and further study of parameter-efficient adaptation in quality assessment.

\section*{Author Contributions}
Bishr Omer Abdelrahman Adam: Conceptualization, Methodology, Software,
Investigation, Formal analysis, Writing -- original draft.
Xu Li: Supervision, Funding acquisition, Writing -- review \& editing.
All authors have read and approved the final version of the manuscript.

\section*{Acknowledgements}
This work is supported by the Key Research and Development Program
of Shaanxi Province (No.~2025CY-YBXM-079).
The authors thank the Kaggle platform for providing free GPU access
used for one-time VLM feature extraction.
The authors also thank the creators of KonIQ-10k, KADID-10k, TID2013,
LIVE-itW, CSIQ, LIVE-MD, and PIPAL for making their datasets publicly
available to the research community.

\section*{Declaration of Competing Interest}
The authors declare that they have no known competing financial
interests or personal relationships that could have appeared to
influence the work reported in this paper.

\section*{Data Availability}
The datasets used in this study are publicly available.
KonIQ-10k is available at
\url{https://database.mmsp-kn.de/koniq-10k-database.html}.
KADID-10k is available at
\url{http://database.mmsp-kn.de/kadid-10k-database.html}.
TID2013 is available at
\url{https://www.ponomarenko.info/tid2013.htm}.
LIVE Challenge in-the-Wild is available at
\url{https://live.ece.utexas.edu/research/ChallengeDB/index.html}.
CSIQ is available at \url{https://s2.smu.edu/~eclarson/csiq.html}.
LIVE Multiply-Distorted is available at
\url{https://live.ece.utexas.edu/research/Quality/live_multidistortedimage.html}.
PIPAL is available at \url{https://github.com/HaomingCai/PIPAL-dataset}.
The source code is publicly available at
\url{https://bishr-omer.github.io/lora-biqa/}.

\end{document}